\documentclass[letterpaper, 10pt, conference]{ieeeconf}      

\IEEEoverridecommandlockouts                              
\usepackage{graphicx} 
\usepackage{amsmath} 
\usepackage{amssymb}  
\usepackage{algorithm}
\usepackage{algorithmic}
\usepackage{xcolor}
\usepackage{amsmath}

\makeatletter
\let\NAT@parse\undefined
\makeatother
\usepackage{hyperref}

\usepackage{float}
\graphicspath{{figures/}}

\title{\LARGE \bf Estimating Motion Codes from Demonstration Videos}

\author{Maxat Alibayev, David Paulius, and Yu Sun
\thanks{Maxat Alibayev, David Paulius and Yu Sun are in the Department of Computer Science \& Engineering at the University of South Florida, Tampa, FL, USA. They are members of the Robot Perception and Action Lab.
\newline(Contact: \texttt{\{alibayevm,davidpaulius,yusun\}@usf.edu)}}
}

\graphicspath{{figures/}}
\usepackage{booktabs}

\begin{document}

\maketitle

\thispagestyle{empty}
\pagestyle{empty}

{
\begin{abstract}
A \emph{motion taxonomy} can encode manipulations as a binary-encoded representation, which we refer to as \emph{motion codes}.  These motion codes innately represent a manipulation action in an embedded space that describes the motion's mechanical features, including contact and trajectory type. The key advantage of using motion codes for embedding is that motions can be more appropriately defined with robotic-relevant features, and their distances can be more reasonably measured using these motion features.
In this paper, we develop a deep learning pipeline to extract motion codes from demonstration videos in an unsupervised manner so that the video' knowledge can be properly represented and used for robots. 
Our evaluations show that motion codes can be extracted from demonstrations of action in the EPIC-KITCHENS dataset. 
\end{abstract}
}


\section{Introduction}

Roboticists have aimed to develop robots or intelligent agents for activities of daily living, that can perform not only tasks for humans but also understand their actions in order to work safely alongside us.
Designing an effective representation of knowledge is very important in drawing meaning from actions or understanding what it has observed or learned \cite{paulius2019survey, jelodar2018long}.
Typically, humans communicate manipulations (such as cutting, mixing, or picking-and-placing, or other everyday household activities) using verbs, but understanding the motions through their physical properties, such as contact types and trajectory types, has not been extensively studied for robots.

The disconnection between motions in a language space for communication and their physical properties poses many difficulties.  First of all, a verb is usually too ambiguous to carry much of the motion's physical features.  A much longer and detailed description, usually with many sentences, would be required to represent a motion clearly.  Second, distances between two verbs in language spaces do not represent their differences in terms of physical motion.  Neither Word2Vec \cite{mikolov2013distributed} nor one-hot representations take physical motion into consideration.  Hence, it is difficult to define loss functions using verbs since they have no physical or mechanical meanings.  Third, the elements in the verb vector either in its original form or Word2Vec have no meanings, making any element-based calculation computation meaningless. These difficulties hinder the way motion knowledge is shared between humans and robots, because eventually, robots rely upon physical features of motions for action. 


For these reasons, we define a new verb embedding called \textit{motion code}, which is based on motion's mechanical attributes, including contact and trajectory features. 
In our previous work \cite{paulius2019manipulation,paulius2020taxonomy}, we introduced a \textit{motion taxonomy}, which is a hierarchical categorization of mechanical properties pertinent to robotic manipulation motions. Using this taxonomy, we illustrated how we could embed the concepts of manipulation as motion codes, which are binary strings or vectors representing motions in a mechanical space. We considered attributes such as contact and trajectory details to be more descriptive of motions such that a robot can then communicate and interact more effectively with humans.
Unlike preceding works such as \cite{kuniyoshi1994learning,dillmann2004teaching,takamatsu2007recognizing,dai2013functional,yang2015robot}, our objective with motion codes is to derive an encoding of motions, which can be used for motion understanding and possibly generation.
Conventionally, motions can be represented as vectors to train models (specifically neural networks) on activity sequences; such vectors can be measured against one another for a variety of tasks, and they have been applied to affordance learning and grounding \cite{huang2017learning_pour,fang2018demo2vec,daruna2019robocse,roesler2019evaluation,tianze2019iros}.
In \cite{paulius2019manipulation}, we introduced characteristics that should be considered when defining motions from the robot's point of view, and we showed how motions translated to their motion codes~\cite{paulius2020taxonomy}. Furthermore, these clusters were supported by a comparison of recorded data from the Daily Interactive Manipulation (DIM) dataset \cite{huang2019dataset,dim}.

\begin{figure}[t]
	\centering
    \includegraphics[trim=0.5cm 2cm 0.5cm 2cm,clip,width=6.7cm]{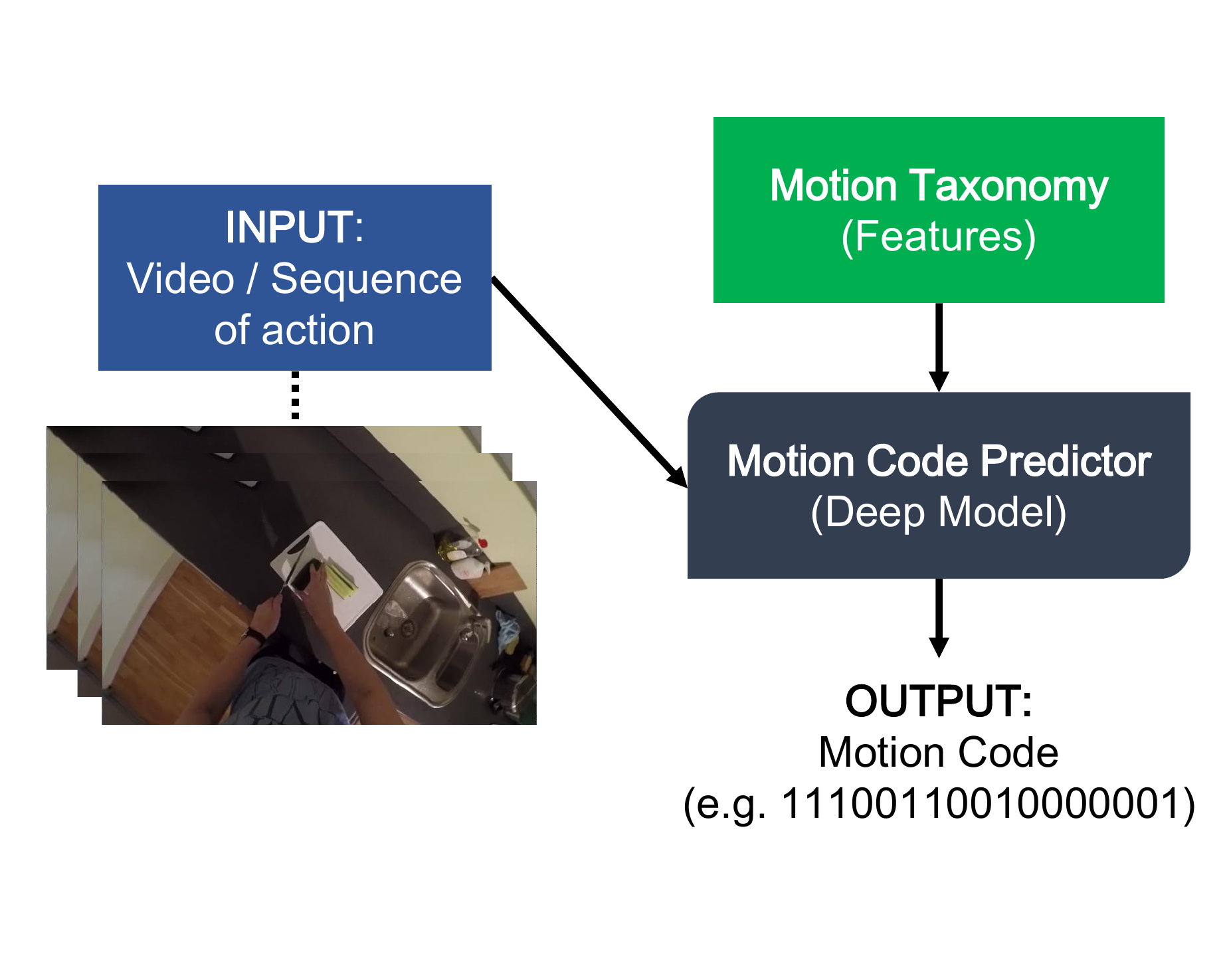}
    \caption{Illustration of the pipeline proposed for motion recognition with the use of the motion taxonomy for motion embedding.
    In this paper, our focus is on \textit{motion code prediction} to obtain a descriptive binary string.}
\end{figure}

However, before this work, motion codes were manually labeled. Making it automatic will allow the robot to understand motions from demonstration videos for learning and collaboration.  Therefore, this paper's objective is to present our approach of extracting motion codes directly from video demonstrations.  This process is referred to as \textit{motion code prediction}, where the goal is to obtain a motion code that accurately describes manipulation from a video. 

First, in the paper, we formally introduce an improved motion taxonomy based on our previous work \cite{paulius2019manipulation} and define its attributes in Section \ref{sec:tax}.
Then, in Section \ref{sec:method}, we discuss a deep-learning-based motion code prediction structure that identifies each attribute of the taxonomy separately as sub-strings that can then be appended together to form a single motion code.  In Section \ref{sec:demos}, we discuss our evaluation of the presented motion code prediction approach and show its accuracy in extracting motion code from test videos{\color{black}, which were obtained from the EPIC-KITCHENS~\cite{Damen2018EPICKITCHENS} dataset}.

\section{Motion Codes}
\label{sec:tax}

In this section, we describe the various attributes used to describe and represent manipulations as motion codes.
The purpose of the motion taxonomy is to translate manipulations into a machine language for motion recognition, analysis, and tentatively generation.
The motion taxonomy from \cite{paulius2019manipulation} was revised and updated in \cite{paulius2020taxonomy}. 
However, in our experiments in Sections \ref{sec:method} and \ref{sec:demos}, we will introduce a condensed version of this taxonomy.
Manipulation is defined to be any atomic action between \textit{active} and \textit{passive} objects; an active object is defined as a tool or utensil (or the combination of a robot's gripper or human hand and tool) that acts upon passive objects, which are objects that are acted upon as a result of motion.
Hence, when annotating demonstrations as motion codes, it is important to identify the active and passive objects in action.


\subsection{Contact Features}
Motion types can be classified as either {\it contact} or {\it non-contact} interaction types.
Contact motion types are all manipulations that require contact between an active object (i.e. the actor's hands or the object that is typically grasped in the actor's hands -- as opposed to the definition in \cite{paulius2019manipulation}) and passive object(s) (i.e. the object(s) that is/are manipulated by the active object) in the work space.
On the other hand, non-contact motion types are all manipulations that do not require contact between active and passive objects or no force is detected between them.
In \cite{paulius2019manipulation}, we also referred to the active object as the \textit{manipulator} and the passive object(s) as the \textit{manipulatee}, but we will instead use the terms active and passive objects in this paper.
Contact can be detected or observed visually (for instance, by the objects' borders or bounding boxes) or using force sensors mounted on objects.
An example of a contact motion is cutting, where the active tool makes contact with a passive object that is deformed into smaller units.
An example of a non-contact motion is pouring; when pouring from one container (active) to another (passive), we usually will not observe contact.

Should a manipulation be identified as contact, we classify the interaction or engagement between objects as either {\it rigid} or {\it soft}.
Rigid engagement is where there is no deformation or structure change among interacting objects, while soft engagement is where objects deform as a result of the interaction (e.g. cutting) or if the objects allow admittance (e.g. piercing).
As a change to the previous version of the taxonomy, we have added a description of the structural integrity of the objects used in order to describe deformation, which is separate to the engagement type and falls under its own tree.
Active and passive objects either undergo no state deformation (non-deforming) or exhibit state or structural change (deforming).
A perfect example of a soft engagement motion is chopping, as an active {\it knife} object will permanently deform the passive object into smaller units; as for a rigid motion, an example of this is picking-and-placing, since the object is simply moved and does not exhibit any structural change.
Deformation can be further distinguished as temporary or permanent, which is attributed to the objects' material or texture.
For instance, squeezing an object such as a sponge will show temporary deformation since it returns to its original shape after the action.
However, in the chopping example from before, this state change is permanent.

%

We also note the duration of contact being made between the active and passive objects.
If the actor only makes contact for a short duration in the manipulation, we consider that contact to be \textit{discontinuous}; however, if the contact between the active tool and passive object is persistent, we consider that contact to be \textit{continuous}.
It is important to note that this perspective changes depending on what is considered to be the active object.
If we consider the actor's hand to be the active tool by itself, then we can assume that once it is grasping a tool for manipulation, there would be continuous contact between the hand and the tool, which is why we consider the active tool to be either the hand (if there is no tool acting upon other objects) or both the hand and tool as a unit (if there are other objects in the manipulation).
Contact duration can be determined visually (for instance, by timing the overlap of bounding boxes) or physically with sensors.


\subsection{Trajectory Features}
An important aspect of manipulation execution is motion planning and generation, where importance lies in executing a trajectory that fulfills the conditions of an action.
Hence, in our taxonomy, we describe the trajectory taken by both the active and passive objects; we describe an object's trajectory as {\it prismatic} (or translational) or {\it revolute} (or rotational).

Prismatic motions are manipulations where an object is translated or moved along a certain axis or plane.
Prismatic motions are identified as 1-dimensional (along a single axis), 2-dimensional (confined to a plane) or 3-dimensional (confined to a manifold space); this can be interpreted as having 1 to 3 degrees of freedom (DOF) of translation.
Revolute motions, on the other hand, are those manipulations where an object is rotated about an axis or plane of rotation; a robot performing such motions would rely on its revolute joints.
Similar to prismatic motions, revolute motions range from 1-D to 3-D motion (i.e. from 1 to 3 DOF); typically, revolute motions are confined to a single axis of rotation in world space (such as in fastening or loosening a screw).

A motion is not limited to one trajectory type, as these properties are not mutually exclusive; therefore, we can say that a motion can be prismatic-only, revolute-only, neither prismatic nor revolute or both prismatic and revolute.
An example of a prismatic-only motion is dipping, where an active object will be translated to make contact with a stationary passive object, while an example of a revolute-only motion is fastening a screw using a screwdriver.
Some motions such as flipping with a turner or spatula may exhibit both prismatic and revolute properties.

We can also describe trajectory by its \textit{recurrence}, which describes if the motion exhibits repetitive movement by repeating its trajectory. 
A motion can be \textit{acyclical} or \textit{cyclical}, which may be useful depending on the context of motion.
For instance, mixing ingredients in a bowl may be repeated until the ingredients have fully blended together, or in the case of loosening a screw, the screwdriver will be rotated until the screw is completely out of the surface.
Learned acyclical motions can be made cyclical by simply repeating the executed trajectory, which is a decision that can be left up to the robot if it is not finished with its task and failed to successfully complete the manipulation.


\begin{figure}[t]
	\centering
    \includegraphics[width=8.5cm,trim={0.7cm .7cm 0.7cm 0.7cm}, clip]{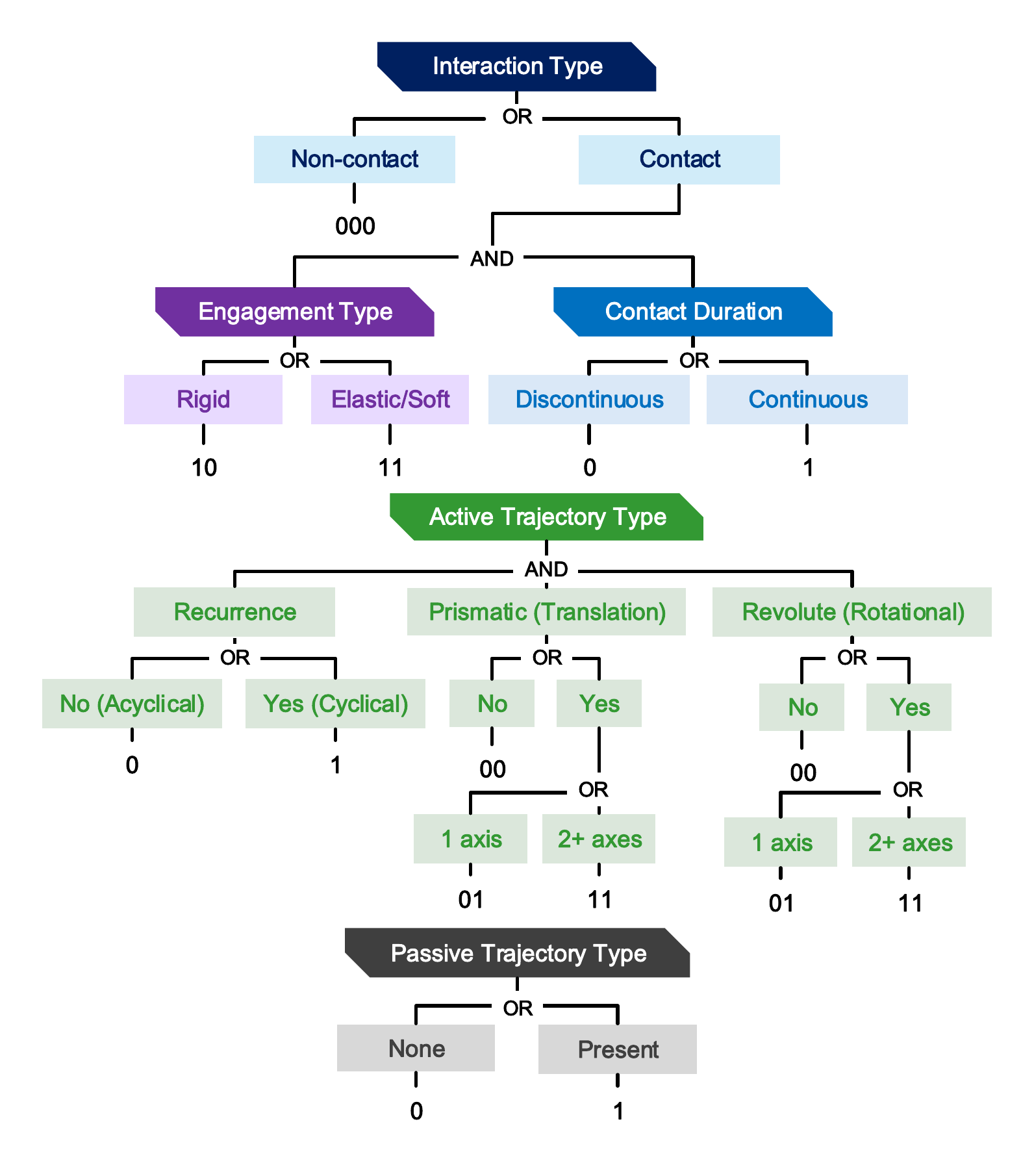}
    \caption{Hierarchy of attributes used in motion code prediction.
    Motion codes were simplified to 9 bits in length, which was done by minimizing the trajectory descriptors for better use on egocentric videos.}
    \label{fig:taxx-mod}
\end{figure}

\subsection{Simplified Taxonomy and Motion Codes}
By default, a full motion code is of length 18 bits~\cite{paulius2020taxonomy}.
However, due to the difficulty of identifying certain features as well as the limitation imposed by our evaluation dataset, we selected only a subset of the attributes for motion codes in this paper, thus shortening the motion code to 9 bits. In particular, we do not consider the {structural outcome} (both active and passive) and {active descriptor} bits, because we wanted to focus more on the interaction between the active and passive objects. In addition, we also simplified the format for trajectory bits; for active object trajectory, we reduced the number of DOF for prismatic and revolute trajectory descriptors to be zero, one, or many (i.e. more than one). We justify this simplification by the fact that we only use visual information from egocentric videos, which makes it challenging to accurately compute the trajectories without a coordinate system of reference.  Lastly, the trajectories of the passive objects were reduced to a single bit representing the existence of the motion of the passive object with respect to the active object, which is different to the default description of passive objects with respect to the world frame. Figure \ref{fig:taxx-mod} shows all applied modifications as a hierarchical tree, which differs to that in \cite{paulius2020taxonomy}.
With 5 unique outcomes for interaction type, 2 unique outcomes for active object's recurrence, 3 outcomes for prismatic and revolute trajectories, and 2 outcomes for passive object's trajectory, the motion codes in this format may have 180 valid combinations.

\begin{figure*}[t]
	\centering
    \includegraphics[trim=0.2cm 2.6cm 0.5cm 6.1cm,clip,width=13cm]{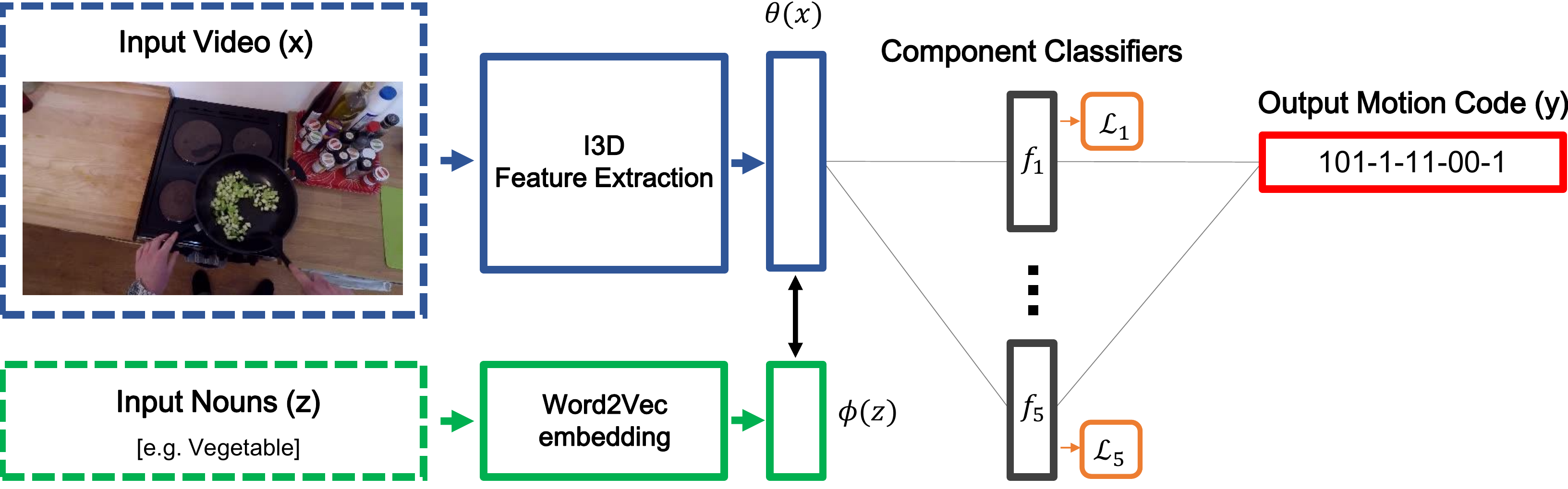}
    \caption{Illustration of the combined network architecture used for motion code prediction. 
    By combining I3D features and Word2Vec embedding with our own classifiers that detect attributes in Figure \ref{fig:taxx-mod}, we output motion codes that describe the manipulation occurring in the input video.
    }
    \label{fig:architecture}
\end{figure*}

\begin{table}[t]
\centering
\caption{Motion codes for manipulations based on Figure \ref{fig:taxx-mod}. Each component is separated by hyphens for legibility.
}
\label{tab:motion_code}
\begin{tabular}{p{2cm}l}
\toprule[1pt]
\textit{\textbf{Motion Code}} & \textit{\textbf{Motion Types and Variants}} \\ 
\midrule[1pt]
\textit{000-0-00-01-0}  &   pour    \\
\textit{000-0-01-00-0}  &   sprinkle \\ 
\textit{{100}-{0-00-01}-{0}} & rotate \\
\textit{{100}-{0-01-00}-{1}} & switch, press, turn on (button) \\
\textit{{100}-{0-11-01}-{1}} & turn on, switch (knob) \\
\textit{{101}-{0-01-00}-{0}} & pull, push, move, wipe, spread \\
\textit{{101}-{0-01-01}-{0}} & turn over, flip \\
\textit{{101}-{0-11-00}-{0}} & put/take (in, out), remove, pick, place \\
\textit{{101}-{0-11-00}-{1}} & open, close (door) \\
\textit{{101}-{1-00-01}-{0}} & shake (revolute) \\
\textit{{101}-{1-01-00}-{0}} & shake (prismatic) \\
\textit{{101}-{1-11-01}-{1}} & shake, wash, move \\
\textit{{110}-{0-11-01}-{0}} & dry, shake (drying) \\
\textit{{111}-{0-00-00}-{0}} & squeeze (in hand) \\
\textit{{111}-{0-01-00}-{0}} & dip, insert, pierce, crack (egg) \\
\textit{{111}-{0-01-00}-{1}} & peel, break, cut, chop, slice, scrape \\
\textit{{111}-{0-01-01}-{0}} & squeeze \\
\textit{{111}-{0-01-01}-{1}} & dry (hands) \\
\textit{{111}-{0-11-00}-{0}} & fold, break, spread, squeeze \\
\textit{{111}-{1-11-00}-{0}} & mix, stir, beat, whisk \\

\bottomrule[1pt]
\end{tabular}
\end{table}

\subsection{Translating Motions to Code}
\label{sec:foon}
Motion codes can be assigned to manipulations by using the flowchart in Figure \ref{fig:taxx-mod} as a decision tree.
We will use the example of flipping a patty with a turner.
First, we start with the interaction type and we determine whether the motion is contact or non-contact.
In flipping, the active turner object makes contact with the passive object, therefore classifying this as a contact motion.
Since there is contact, we proceed down that branch, where we then describe the engagement type between the objects and the contact duration throughout the
action.
In our example, the turner will lift and not deform the passive object while maintaining contact for a majority of the manipulation (until the passive object falls). Hence, for the contact portion of the code, we will have \textit{`101'}.
We then proceed to describe the trajectory type of the objects in action.
When flipping an object, there is usually no recurrence (which we indicate with \textit{`0'} before the trajectory bits).
The active trajectory has some 1D prismatic trajectory and it especially exhibits rotation about a single axis (\textit{`0-01-01'}); the passive object also adopts the same trajectory, therefore making no movement relative to the active object (\textit{`0'}).
By combining all of these substrings, we end up with a single, representative motion code \textit{`101-0-01-01-0'}.

In Table \ref{tab:motion_code}, we provide examples of action types seen in several household manipulation datasets, such as DIM, EPIC-KITCHENS, MPII Cooking Activities~\cite{MaxPlankIICooking}, and FOON~\cite{Paulius2016,Paulius2018}, and their respective motion code assignment.
Several motions can share the same motion code due to common mechanics, such as cutting and peeling since they are both 1D-prismatic motions that permanently deform the passive objects.
We can also account for variations in manipulation.
For instance, we may see manipulations that operate in 1 DOF or in many DOF, or we may see different motions with the same label (as we will show in Section \ref{sec:demos}).



\section{Methodology}
\label{sec:method}
Having established the definition of the motion taxonomy and all properties it encompasses, we now discuss how we can automatically obtain motion codes from video demonstrations. In short, we design a deep neural network model to extract visual feature vectors from the videos, which are then further passed to classifiers for each component of the taxonomy.
The overall structure is illustrated as Figure \ref{fig:architecture}.

\subsection{Motion Prediction Model}
Given a video $x$ (where $x \in X$), we want to obtain a motion code $y$ that describes the manipulation taking place by feeding $x$ into a deep network $\theta: X \to \Omega$.
Since our goal is to use video modality for motion code prediction, the deep network model $\theta$ should integrate spatial and temporal features of the videos. 
Presently, there are several action recognition models that satisfy these criteria.
Our choice fell on Two-Stream Inflated 3D ConvNets (I3D), originally from \cite{carreira2017quo}.
This architecture uses Inception-V1 \cite{Szegedy_2015_CVPR} convolutional neural network (CNN), which was pre-trained on ImageNet dataset, and inflates the 2D convolutional and pooling layers with a temporal dimension that were later tuned with the Kinetics dataset \cite{carreira2017quo}. 
The model also combines two modalities of video frames: RGB and optical flow frames. 
Two separate models for both modalities are trained individually, and the final feature vectors are obtained via late fusion by averaging two outputs. 
This model boasts the highest action recognition accuracy results on well-known benchmarks, including EPIC-KITCHENS \cite{Damen2018EPICKITCHENS}, which we use in our experiments. 
To potentially improve motion code prediction, we also tried to incorporate knowledge of the objects in action into the training process.
Formally, we modified the model described above by encoding the semantic features of the objects $z \in Z$ with embedding function $ \phi: Z \to \Psi $ and combining it with the visual features.
For our experiments, we concatenate these two feature vectors.
We use a Word2Vec model pre-trained on Google News \cite{mikolov2013distributed} (containing over 3 million words) to encode these semantic features about objects seen in each video.
A model of this kind could be used for queries, where we can determine what kind of motion with a given object can be executed to replicate a certain activity. 

From each video, we use the extracted features $\omega$ from I3D (where $\omega \in \Omega$) and the word embedding of object semantics $\psi$ (where $\psi \in \Psi$) as a feature vector $\xi \in (\Omega,\Psi^{*})$ 
(where $\Psi^{*}$ denotes that semantic features may or may not be used)
to be passed into several classifiers $f_{i}$.
To explain, each motion code was broken down into the five independent components: interaction type, recurrence, active prismatic motion, active revolute motion, and passive object motion with respect to active object. 
Each of these components were used to train individual classifiers whose output will be combined into a single code as opposed to predicting an entire motion code.
In this way, individual predictors will more accurately identify features.
In other words, the feature vector is then passed through these classifiers $f_{i}: (\Omega, \Psi^{*}) \to Y_{i}$ to then predict the value for the $i^{th}$ component, $y_{i} \in Y_{i}$, of the final motion code.
Each component was converted into one-hot vectors, making the motion prediction identical to feature classification. 
All components were then combined into a single motion code $y$.
The objective function for classification of the $i^{th}$ component, $\mathcal{L}_{i}$, is a cross-entropy loss. The total objective function is the linear combination of all components' objective functions, as follows:
\begin{equation*}
    \mathcal{L} = {\sum}_{i=1}^{5} \lambda_{i}\mathcal{L}_{i} \\[1pt]
\end{equation*}

\section{Evaluation}
\label{sec:demos}
Based on the proposed methodology, we now evaluate the performance of unsupervised motion code prediction from video demonstrations using our proposed model.
Using videos from EPIC-KITCHENS, we evaluated three variations of our motion prediction model and show how well they performed in deriving motion codes.

\subsection{Dataset and Training Details}
For our motion prediction model, we use videos from the EPIC-KITCHENS~\cite{Damen2018EPICKITCHENS} dataset for training.  We annotated 3,528 video segments with motion codes using the taxonomy (Figure \ref{fig:taxx-mod}).  Videos in EPIC-KITCHENS are annotated with object in action details in the ``noun'' field, where the entry is a list of words that describes the objects. We use these words as an input $z$ to the Word2Vec model. Overall, 2,742 videos were used for training and 786 were used for testing.  The dataset was annotated with 32 unique motion codes, and each code had at least 20 video segments assigned as its label.

The entire model was trained for 50 epochs with Adam optimizer and the learning rate set to 0.0003 that decreases by 40\% every 5 epochs. For the first 3 epochs, the convolutional layers of the base model were frozen to allow the top layers to fine-tune for a better initialization.
The input video frames are sampled to 6 frames per second to increase the training and inference speed. 
As mentioned before, both RGB and optical flow frames were used. 
In our experiments, all $\lambda$ coefficients are set to $1$. During training, we add the $L_{2}$ norm of the network parameters multiplied by a weight decay factor to the loss function for $L_{2}$ regularization. 
The model was implemented with the TensorFlow library.

\subsection{Results and Analysis}
We computed the results for three variations of our model: one that uses only visual features, another that incorporates noun vectors with visual features, and another that also uses noun vectors but with noise, where 20\% of nouns are incorrect. 
For each model, we also evaluated the performance of each individual component of the motion codes, namely the interaction (3 bits), recurrence (1 bit), active prismatic motion (2 bits), active revolute motion (2 bits), and passive object motion with respect to the active object (1 bit). 
The models were evaluated based on their motion code prediction accuracy. 
From Table \ref{tab:resuts} we can see that the baseline model performed fairly well given that there are not many training videos and 180 possible outcomes. 
However, when we relaxed the accuracy measurement to allow at most one bit to be incorrectly predicted, the overall performance increased by 32\%, which means roughly a third of all predicted codes are just one bit off from the ground truth motion code. 
Another interesting thing to note is that in all cases, late fusion of the predictions of both modalities improves the overall performance of the model.

\begin{table}[t]
\centering
\caption{Motion prediction accuracy results on test videos (as \%).}
\label{tab:resuts}
\begin{tabular}{llll}
\toprule[1pt]
\textit{\textbf{Models}} & \textit{\textbf{RGB}} & \textit{\textbf{Flow}} & \textit{\textbf{Fused}}\\ 
\midrule[1pt]
\textit{\textbf{Baseline}} & & & \\
Entire code & 35.1 & 35.2 & 38.9 \\
Entire code with 1 bit off & 67.3 & 64.5 & 70.9 \\
\\
Interaction & 85.8 & 84.7 & 87.0 \\
Recurrence & 90.7 & 91.0 & 92.5 \\
Prismatic trajectory & 70.6 & 72.8 & 73.2 \\
Revolute trajectory & 74.4 & 76.2 & 78.5 \\
Passive motion & 68.6 & 64.8 & 71.9 \\

\midrule[1pt]
\textit{\textbf{Nouns}} & & & \\
Entire code & 45.3 & 46.1 & 48.0 \\
Entire code with 1 bit off & 73.2 & 72.1 & 75.3 \\
\\
Interaction & 86.4 & 86.4 & 87.9 \\
Recurrence & 90.6 & 91.2 & 92.1 \\
Prismatic trajectory & 76.0 & 74.4 & 74.9 \\
Revolute trajectory & 80.5 & 78.6 & 81.3 \\
Passive motion & 76.0 & 78.9 & 79.9 \\

\midrule[1pt]
\textit{\textbf{Nouns (20\% noise)}} & & & \\

Entire code & 40.8 & 39.9 & 43.1\\
Entire code with 1 bit off & 70.7 & 68.8 & 72.1 \\
\\
Interaction & 86.5 & 85.9 & 88.0 \\
Recurrence & 90.7 & 91.5 & 92.5 \\
Prismatic trajectory & 74.2 & 71.8 & 73.5 \\
Revolute trajectory & 76.3 & 76.6 & 78.4 \\
Passive motion & 72.4 & 72.6 & 73.9 \\
\bottomrule[1pt]
\end{tabular}
\end{table}

Results in Table \ref{tab:resuts} also suggest that using noun vectors as an input to the model improves the overall accuracy by ~10\%, which is quite a significant jump.
If we look at the breakdown of individual components, noun vectors mostly benefit the prediction of passive object motion with respect to the active object. 
We observed that roughly 10\% of videos that were assigned incorrect motion codes by the baseline model but were then given correct codes from the noun model had only 1 bit that was wrong, which was the passive object motion bit. 
For instance, the baseline model classified a video as passive object motion being present with respect to the manipulator, while the ground truth is the opposite.

Interestingly, 8\% of these videos showed an action where person either picks up and places an object or opens and closes it (e.g. door, microwave, fridge). The passive object motion bit for pick-and-place actions were corrected from 1 to 0, while in open and close actions, the bit was set from 0 to 1. In almost all cases when a person picks or places an object, the object is moving with the same trajectory as the hand, making the passive object stationary with respect to the hand. 
On the other side, when person opens or closes a door, the door always moves strictly around its axis, making it to have only a revolute motion with 1 DOF. 
Such kind of a motion trajectory is rarely performed by a human. These examples show that the model leverages the information about the objects and can make assumptions based on that knowledge.
Even after adding 20\% noise to the input noun vectors, the overall accuracy of motion code prediction lies right in the middle between the baseline model and the model with 100\% correct noun vectors. 

{
Overall, our model shows that it can successfully predict motion components. The performance can be further improved with a larger training set due to the large number of parameters to train in the I3D model. The EPIC-KITCHENS dataset includes 28,472 training videos, which is ten times larger than the size of our training set, but labelling these videos with ground truth motion codes is time-intensive. We would also want to be able to predict the motion codes of the original length and format that would define the motions in a more detailed way. However, the information from egocentric videos is not sufficient for this task. This could be achieved with ADL videos of the same scenes from multiple perspectives. In addition, readings from motion sensors could be helpful in predicting motion trajectories of all objects with respect to a fixed point in the world coordinate system.
}

\section{Conclusion}
\label{sec:con}
In this paper, we build upon verb embedding as \textit{motion codes}, which we proposed for a robot's understanding of motions in a physical space~\cite{paulius2019manipulation,paulius2020taxonomy}. 
Motion codes are created using the \textit{motion taxonomy} to describe motions from a mechanical perspective.
Prior to this work, motion codes were not obtained automatically from demonstration videos; therefore, in this work, we introduced our approach for extracting motion codes directly from video demonstrations, which we refer to as \textit{motion code prediction}.
Using existing neural network models, we designed a deep model that identifies components from the taxonomy separately, which are then concatenated to form a single motion code.

Our experiments showed an accuracy of 70.9\% when we allowed at most 1 bit to be incorrect from the ground truth using our baseline model that only considers visual features; however, by integrating nouns associated with each video with the model, we showed that we can improve this performance to 75.3\%.
Even when considering an overall correct prediction, noun integration significantly improved results by ~10\%.
The achieved results lead us to believe that embedding motions with our taxonomy may also contribute to action recognition from videos.
Such an approach to embedding introduces a new motion space that can bridge the gap between visual features and the action labels and other semantic features. In addition, the generated motion codes could be used as an attribute space for zero shot learning.

In the future, we will investigate if motion codes can improve accuracy in motion recognition tasks that use human language labels for supervised learning. Additionally, with an improved motion prediction model, we will explore the use of motion codes to append knowledge and to perform task planning with our knowledge source FOON~\cite{Paulius2016,Paulius2018}. 

\section{Acknowledgements}

\noindent This material is based upon work supported by the National Science Foundation under Grant Nos. 1812933 and 1910040.

\bibliographystyle{IEEEtran}
\bibliography{references}

\end{document}